\documentclass[a4paper,twocolumn,twoside,10pt]{ranlp}

\begin{document}

\title{\textbf{Generating models for temporal representations}}

\author{
Patrick Blackburn\\
INRIA Lorraine\\
615 rue du Jardin Botanique\\
54602 Villers lès Nancy Cedex, France\\
\email{Patrick.Blackburn}{loria.fr}
\and
S\'ebastien Hinderer\\
INRIA Lorraine\\
615 rue du Jardin Botanique\\
54602 Villers lès Nancy Cedex, France\\
\email{Sebastien.Hinderer}{loria.fr}
}

\date{}
\maketitle
\begin{abstract}

We discuss the use of model building for temporal representations.  We
 chose Polish to illustrate our discussion because it has an
 interesting aspectual system, but the points we wish to make are not
 language specific.  Rather,  our goal is to develop theoretical and
 computational tools for temporal model building tasks in computational
 semantics. To this end, we present a first-order theory of time and
 events which is rich enough to capture interesting semantic
 distinctions, and an algorithm which takes minimal models for
 first-order theories and systematically attempts to ``perturb''
 their temporal component to provide non-minimal, but semantically
 significant, models.

\end{abstract}


\section{Introduction}

In this paper we discuss the use of model building for temporal
representations.  We chose Polish to illustrate the main points
because (in common with other Slavic languages) it has an interesting
aspectual system, but the main ideas are not language specific.
Rather, our goal is to provide theoretical and computational tools
for temporal model building tasks. To this end, we present a
first-order theory of time and events which is rich enough to capture
interesting semantic distinctions, and an algorithm which takes
minimal models for first-order theories and 
systematically attempts to ``perturb'' their temporal component to
provide non-minimal, but semantically significant, models.

The work has been implemented in a modified version of the \textsf{Curt}
architecture.  This architecture was developed by Blackburn and
Bos~\cite{BlackburnBos:2005} to illustrate the interplay of logical
techniques useful in computational semantics. Roughly speaking, the
\textsf{Curt} architecture consists of a representation component
(which implements key ideas of Montague semantics~\cite{monty}) and an
inference component. In this paper we have used a modified version of
the representation component (based on an external tool called
\textsf{Nessie} written by S\'ebastien Hinderer) which enables us to
specify temporal representations using a higher-order logic called
$TY_4$. However, although we shall briefly discuss how we build our
temporal representations, the main focus of the paper is on the
other half of the \textsf{Curt} architecture, namely the inference component.

Inference is often though of simply as theorem proving.  However one
of the main points made in \cite{BlackburnBos:2005} is that a wider
perspective is needed: theorem proving should be systematically
coupled with model building and the \textsf{Curt} architecture does
this.  Model building takes a logical representation of a sentence and
attempts to build a model for it; to put it informally, it attempts to
return a simple picture of the world in which that formula is
true. This has a number of uses. For example, as is emphasized in
\cite{BlackburnBos:2005}, model building provides a useful positive
test for consistency; if a model for a sentence can be built, then
that sentence is consistent (this can be useful to know, as it enables
us to prevent a theorem prover fruitlessly searching for a proof of
inconsistency).  Moreover, in subsequent papers, Johan Bos and his
co-workers have demonstrated that model building can be a practical
tool in various applications (see for example \cite{bos1,bos2,bos3}).

The work described here attempts to develop a \textsf{Curt} style
architecture rich enough to handle natural language temporal
phenomena. So far we have concentrated on the semantic problems raised by
tense and aspect. We have developed a first-order theory of time and
events, which draws on ideas from both \cite{mas} and
\cite{baf}. Although these theories were developed for English, we
believe the underlying ideas are more general, and to lend support to
this claim we shall work here with Polish.

As we shall see, however, more than a theory of time and events is
required. Model builders typically build the smallest models possible,
but such models may not be suitable for all tense and aspectual
combinations, which often underspecify the temporal profile of the
situations of interest. We thus provide an algorithm which takes as
input a first-order theory, a first-order formula, and a model for the
theory and formula, and systematically attempts to ``perturb'' the
temporal part of the model to find non-minimal but semantically
relevant models.

\section{Modelling tense and aspect}

In this section, we shall discuss the logical modeling of tense and
aspect, drawing on some simple examples from Polish, and   informally
introduce a temporal ontology of time and events which will let us
express temporal and aspectual distinctions in a precise way.  The
formal definition of a theory over this temporal ontology (which draws
on ideas from \cite{baf} and \cite{mas}) will be given in
Section~\ref{theTheory}.

Consider the following four Polish sentences:

\begin{enumerate}
\item Piotr pospaceruje 
\item Piotr pokochal Aline 
\item Piotr napisal list \\
\mbox{   } \ \textit{and} \\
      Piotr popisal list 
\end{enumerate}

The first sentence refers to a walking event and adopts a perfective
point of view: it insists on the fact that the mentioned action will
be terminated at some point in the future.  The second sentence
mentions an eventuality of loving and also adopts a perfective point
of view. However, the reading of this sentence differs from the
previous one. The first sentence insisted on the {\em termination} of
the event, whereas the second one insists on its {\em beginning}. In
other words, the second sentence has an {\em inchoative} reading. This
is because the verb ``kocha'' from which ``pokochac'' is derived is a
{\em state verb}, and perfective state verbs have inchoative readings
in Polish.  So the second sentence means that at some point in the
past Piotr started to love Alina.

The last two sentences, which are also perfective, both refer to the
termination of a writing event which is located in the past. The
difference between these two sentences concerns the way the writing
event terminated. In the ``napisac'' variant, an idea of successful
termination is conveyed: that is, at some point the writing stopped,
because the letter was finished. In the ``popisal'' variant, the
writing also stopped but the conveyed idea is that the writing event
was interrupted before its normal termination, which implies that the
letter could not be finished.  To distinguish between a
``normal'' termination and a termination due to an unexpected,
premature interruption, we talk about {\em culminations}.  An event
{\em culminates} when it terminates and has also been completed, or fully
accomplished. Thus the event of writing reported by the sentence
``Piort napisal list'' culminates, whereas the one in ``Piotr popisal list''
does not.

Note that in our two first examples, it makes no sense to talk about
the culmination of the walking or loving eventualities; neither
walking events nor states of loving have natural culminations in the
way that writing events do.  More generally, different types of events
have different properties, and verbs can be classified according to
the properties of the event they refer to. Such a classification has
been proposed for Polish verbs by M{\l}ynarczyk~\cite{mlyn:aspe04},
and we follow this classification in our work.  The classification
proposes five verb classes, including the three just mentioned: a
class for processes (``to walk'' belongs to this class), a class of
state verbs and gradual transitions (a member of which is ``to love'')
and a class for culminations (``to write'' belongs to this class).
Processes are non-instantaneous events which have no particular
properties; it is possible to look at them either as ongoing
(imperfective), or as finished (perfective). State verbs are also non
instantaneous. Their imperfective use corresponds to a vision of the
state as holding, whereas (as was already mentioned) their perfective
use has an inchoative reading. Culminations have an imperfective
variant and two perfective ones: one for events that have culminated,
another for event that have not culminated.

Now, our aim is to translate simple Polish sentences like those just
discussed into logical formulas that encode their meaning. More
precisely, we are interested in obtaining logical formulas that give
an account of the sentence's temporal and aspectual properties
suitable for theorem proving and model building purposes.  This means
we should choose a logic that makes it easy to distinguish various
kinds of entities (for example, ordinary individuals and events) and
that lends itself naturally to semantic construction.  To achieve
these goals we will use a higher-order typed logic called $TY_4$.
This logic belongs to the $TY_n$ family of logics.  This family of
logics has long been advocated by Muskens (see, for example,
\cite{musk:mean96}) as an appropriate logical setting for natural
language semantics.  The four basic vocabulary types we shall build
the formulas of this logic over (in addition to the type of
truth-values which is always included in $TY_n$ theories) are:
\begin{description}
\item[entity]: for  individuals and objects;
\item[time]: for  moments of time;
\item[event]: for the  events introduced by  verbs;
\item[kind]: to classify  events into kinds.
\end{description}\par

The first type (entity) will certainly be familiar to the reader used
to Montague-style semantic construction. The second type, time, is
clearly needed to give an account of notions like past, present and
future.  The abstract entities known as events (introduced by
\cite{Davidson}) are a convenient object one can use to talk about
actions introduced by verbs.  Each verb introduces an event, which is
then used to record additional information about the action the verb
describes. For example, if the verb ``to eat'' introduces an event
$e$, then the fact that the entity doing the eating is $x$ will be
encoded as ${\textsc agent}(e,x)$, the fact that the eaten entity is $y$
will be encoded as ${\textsc patient}(e,y)$, and so on. Event-based
representations for the verbs make it easy to attach
additional information, for example information contributed by verb
modifiers; for each modifier, one simply introduces a binary predicate
whose first argument is the event of interest and whose second
argument is the piece of information to be attached to this event.
Finally, every event has a kind, and we assume that each verb picks
out a distinct kind of event.

The logic we work with makes use of the following binary predicates
relating events and times:

\begin{itemize}
\item ${\emph inception}(e,t)$ means that the event $e$ starts to take
  place at the moment $t$;
\item ${\emph conc}(e,t)$ means that the event $e$ ends at the
  moment $t$;
\item ${\textsc induration}(e,t)$ means that the event $e$ is going on at
    the moment $t$;
\item ${\textsc ek}(e,k)$ means that the event $e$ is of kind $k$.
\end{itemize}

In addition, it has the following binary relation which relates times:

\begin{itemize}
\item ${\emph lt}(t,t')$ means that  time $t$ 
  is before time  $t'$.
\end{itemize}

Furthermore, it has the following   binary relation between events:
\begin{itemize}
\item ${\emph culm}(e,e')$ means that  event $e'$ 
  is the culmination of event   $e$.
\end{itemize}
This relation  plays a key role in analysing the semantics of verbs
like ``napisal/popisal''.

There are also number of other unary relations involving events (such
as \textit{culminated}($e$)), and a temporal constant {\textsc now} to
represent the time of utterance.  The way these items are inter-related
will be formally spelt out in Section~\ref{theTheory}.

\section{Computing semantic representations}

Before turning to the formal specification of the theory of time and
events, we shall briefly outline the process that allows us to
automatically translate Polish sentences into a logical formula over
the vocabulary introduced in the previous section.  This process is
done in three steps: parsing, computing a semantic representation in
higher-order logic, and translating this representation to plain
first-order logic. 
The translation process uses a modified
version of the \textsf{Curt} architecture.

\subsection{Parsing}

The parsing is done using a Prolog DCG. It parses the text given as
input and produces a syntax tree reflecting its structure. The leaves of
this tree can be labelled either by a word and its syntactic category,
or by an operator encoding a verb's temporal and aspectual meaning.

For example, here is the parse tree produced for the sentence ``Piotr
pospaceruje'' (Piotr will have walked):

\begin{verbatim}
binary(s,
       unary(np, leaf(piotr, pn)),
       binary(vp, leaf(pastiv, op), 
              leaf(pospacerowac, iv))
      )
\end{verbatim}

The first and third leaves refer to lexical entries, whereas the
second carries an operator. This operator indicates that the verb
carried by the following leaf is in the past.

\subsection{Computing higher-order logic representations}

This step is performed by an external tool that has been especially
developed to compute semantic representations from a parse tree.  The
tool is called \textsf{Nessie}, and it takes as input a parse tree
similar to the one just presented and a lexicon specifying the
semantic representation for each word; it was  designed to
handle the $TY_n$ family of logics.  Thus for present purposes we
simply declare to \textsf{Nessie} the four basic vocabulary types we
have selected (namely entity, time, event, and kind) and
\textsf{Nessie} is then equipped to handle the higher-order language
they give rise to.  The simply typed lambda-calculus lies at the
heart of the $TY_n$ family of logics, and \textsf{Nessie} handles such
tasks as type-checking and $\beta$-reduction.  In other words, the
work \textsf{Nessie} does is very much inspired by Musken's adaptation
of Montague's original approach to natural language semantics.

The output of this second step of processing is, generally speaking,
a typed lambda-term.  In our temporal representations, once
\textsf{Nessie} has $\beta$-reduced the term, there will be neither
applications nor abstractions present in the final formula. In other
words, the semantic formula provided by this second step is close to a
genuine first-order formula, the only difference being that the
variables occurring in the term are typed.

To continue with our example, \textsf{Nessie} would compute the
following representation for the sentence:%

\medskip
$
\exists t : {\textsc time}. \exists e : {\textsc event}.
( {\textsc lt}({\textsc now},t) \wedge {\textsc ek}(e,{\textsc spacerowac}) \\
\mbox{  }
\ \ \ \ \ \ \ \ \ \ \ \ \ \ \ \ \ \ \  \ \ \ \ \ \
\wedge {\textsc agent}(e, {\textsc piotr}) \wedge {\textsc conc}(e,t) ).
$

\subsection{From higher-order to first-order representations}

In logical semantics there are important
trade-offs between higher-order and first-order logics.  As Montague,
Muskens and others have demonstrated, higher-order logics are a
natural medium for specifying semantic theories: their expressivity
allows semantic representations for all syntactic categories to be
given (and entailment relations between them to be stated).  Moreover,
the fact that they incorporate the simply typed lambda calculus gives
a uniform and simple approach to semantic construction.

But higher-order approaches have a drawback. They are inherently more
complex than first-order approaches. Because of this, relatively few
automated reasoning tools exist for higher-order logics, and those that
do are not particularly efficient. But all is not lost.  As formal
semanticists have long known, in natural language semantics, the
higher-order constructs typically produce representations which are
very close to first-order ones. So, if we could translate the $TY_n$
expressions output by \textsf{Nessie} into first-order logic, we could
have the best of both worlds.

At first glance, it could seem that the only thing to do to convert a
higher-order formula (like the one shown above) into a first-order
one is to remove the types. In fact, things are slightly more
complex than this, as the following example should make
clear. Consider the formula: $\Phi=\forall X : \tau P(x)$, where
$\tau$ is a type. If we throw types away too quickly, we get as
candidate for a first-order translation of $\Phi$: $\Phi'=\forall X
P(X)$.  But $\Phi$ and $\Phi'$ don't have the same meaning: the former
formula states that the predicate $P$ holds for every object of type
$\tau$, whereas the latter claims that $P$ holds for every object, no
matter what its type is.

A semantically correct translation can however be obtained, with the help
of a unary predicate that characterizes the object of type $\tau$. With
the help of such a predicate (which will be written $\tau'$), it becomes
possible to propose a semantically correct translation of $\Phi$ in
first-order logic:
$\Phi''=\forall X (\tau'(X) \rightarrow P(X))$.
To obtain a complete specification of a translation function translating
higher-order formulas into firs-order formulas, a similar trick should
be used for the existential quantifier: $\exists X : \tau P(X)$ is
translated to $\exists X (\tau'(X) \wedge P(X))$.
The translation of other formulas is straightforward.

The complete translation mechanism has been implemented in
\textsf{Nessie} which can on demand produce either higher-order or
first-order semantic representations. Thus, here is the final
first-order representation we get for our initial sentence:

\medskip
$
\exists t 
(\textsc{time}(t) 
\wedge \exists e 
(\textsc{event}(e) \\
\mbox{  } \ \ \ \ \ \ \ \ \ \
\wedge 
\textsc{lt}(\textsc{now},t) \wedge \textsc{ek}(e,\textsc{spacerowac})\\
\mbox{  } \ \ \ \ \ \ \ \ \ \
\wedge \textsc{agent}(e, \textsc{piotr}) \wedge \textsc{conc}(e,t) )).
$

\section{A first-order theory of time and events}
\label{theTheory}

We are interested in computationally modeling tense and
aspectual distinctions. In particular, we want to derive logical
representations useful for model building purposes.  But we have not
yet achieved this goal. Although \textsf{Nessie} can output
first-order representations, simply giving such representations to a
first-order model builder won't give us what we want, for as yet we
have said nothing about how the various symbols we are using are
interrelated.  For example, the previous representation talks about an
event taking place in the future, as the $\textsc{lt}(\textsc{now},t)$
conjunct makes clear. A model for such a representation should of
course reflect this. But nothing in the representation itself prevents
the model builder from identifying $t$ with \textsc{now}, or from
building a model where both $\textit{now}< t$ and $t < \textit{now}$
hold, as we have said nothing about the properties of \textsc{now} or
\textsc{lt} or how they are related.  And this is only the tip of the
iceberg. It is relatively clear what properties \textsc{lt} should
have (for example, it should be transitive) but many other constraints
(notably on the way times and events are interrelated) need to
be expressed too.  In short: to automatically compute models for a
semantic representation, we need to work with respect to a theory of
time and events, and the purpose of this section is to sketch the
theory we use.

In essence, the theory we need should take into account some basic
typing facts (for example that two objects of different types can not
be identified, and that predicates impose typing constraints over
their arguments), structural properties of time (such as the
transitivity of \textsc{lt}), and, most importantly of all, the way
times and events are inter-related.  The following sections give
first-order axioms which formalise the required constraints.  We won't
give all the axioms (for example, we omit all axioms covering events
for verb classes not discussed here) but we have given enough to
convey a flavour of what is required to carry out model building for
tense and aspectual information.

\subsection{Type definitions}

The following axioms state that the set of elements of the models should
be partitioned by the four types we use: event, kind, time and entity.
The following two axioms are typical:

not\_event\_entity : $\forall$A$\neg$(\textsc{event}(A) $\land$ \textsc{entity}(A)) \\

not\_entity\_time : $\forall$A$\neg$(\textsc{entity}(A) $\land$ \textsc{time}(A)) \\





There is also an  axiom stating that every object should belong to at least one type.

\subsection{Typing constraints}

Another family of axioms reflects the typing constraints imposed by the
predicates over their arguments. For example, the binary predicate
\textsc{agent} requires that its first argument is an event and that
its second argument is an entity. The following is a sample
of such axioms:

\medskip


now\_type: \\
\textsc{time}(now) \\

lt\_type: \\
$\forall$A$\forall$B(\textsc{lt}(A,B) $\to$ (\textsc{time}(A) $\land$ \textsc{time}(B))) \\

agent\_type: \\
$\forall$A$\forall$B(\textsc{agent}(A,B) $\to$ (\textsc{event}(A) $\land$ \textsc{entity}(B))) \\


conc\_type: \\
$\forall$A$\forall$B(\textsc{conc}(A,B) $\to$ (\textsc{event}(A) $\land$ \textsc{time}(B))) \\


inception\_type: \\
$\forall$A$\forall$B(\textsc{inception}(A,B) $\to$ (\textsc{event}(A) $\land$ \textsc{time}(B))) \\




ek\_type: \\
$\forall$A$\forall$B(\textsc{ek}(A,B) $\to$ (\textsc{event}(A) $\land$ \textsc{kind}(B))) \\

\subsection{Structure of time}

The previous two groups of axioms were essentially organisational:
they laid out the basic constraints individuating types and imposed
restrictions and requirements on the relations the various types of
entity could enter into. We are now ready to turn to more substantial
axioms, that is, axioms that impose structure on our ontology.  The
simplest such axioms are those  regulating the temporal part of the
ontology.  The following requirements are  standard (see for
example \cite{vb1}):

\medskip

lt\_irreflexive: \\
$\forall$A$\neg$\textsc{lt}(A,A) \\

lt\_transitive: \\
$\forall$A$\forall$B$\forall$C((\textsc{lt}(A,B) $\land$ \textsc{lt}(B,C)) $\to$ \textsc{lt}(A,C)) \\

lt\_total: \\
$\forall$A$\forall$B((\textsc{time}(A) $\land$ \textsc{time}(B)) $\to$ (\textsc{lt}(A,B) $\lor$ (\textsc{eq}(A,B) $\lor$ \textsc{lt}(B,A)))) \\

Other axioms could be imposed (such as the requirement that every
point has a successor, or that the structure of time is dense) but for
present purposes we won't make use of such options.  Instead we will
turn to the heart of our formalisation, namely its treatment of events
and the way they interact with time.  This part draws on and
generalises ideas presented in \cite{baf} and \cite{mas}.

\subsection{Structure of events}

This group of axioms
is itself divided into three parts, namely general
axioms regulating the relationship between times and events, axioms
for instantaneous events, and axioms for culminations (actually, in the
full version of the theory there are axioms constraining the events
required for other verb classes, but we omit them here).

\subsubsection{Relating times and events}

The following is a sample of the axioms we use to regulate the
interplay between the structure of time and the structure of
events. As a rough mental picture, it may be useful to think of events
as hanging from the temporal structure (a bit like balloons hanging by
string from a long stick). The following axioms (which have been
abstracted from \cite{baf}) then ensure that the two kinds of entity
are properly coordinated:

\medskip

agent\_unique: \\
$\forall$A$\forall$B$\forall$C((\textsc{agent}(A,B) $\land$ \textsc{agent}(A,C)) $\to$ \textsc{eq}(B,C)) \\


event\_has\_inception: \\
$\forall$A(\textsc{event}(A) $\to$ $\exists$B\textsc{inception}(A,B)) \\

inception\_unique: \\
$\forall$A$\forall$B$\forall$C((\textsc{inception}(A,B) $\land$ \textsc{inception}(A,C)) $\to$ \textsc{eq}(B,C)) \\

event\_has\_conc: \\
$\forall$A(\textsc{event}(A) $\to$ $\exists$B\textsc{conc}(A,B)) \\

conc\_unique: \\
$\forall$A$\forall$B$\forall$C((\textsc{conc}(A,B) $\land$ \textsc{conc}(A,C)) $\to$ \textsc{eq}(B,C)) \\

inception\_not\_after\_conc: \\
$\forall$A$\forall$B$\forall$C((\textsc{inception}(A,B) $\land$ \textsc{conc}(A,C)) $\to$ $\neg$\textsc{lt}(C,B)) \\

duration\_before\_conc: \\
$\forall$A$\forall$B$\forall$C((\textsc{induration}(A,B) $\land$ \textsc{conc}(A,C)) $\to$ \textsc{lt}(B,C)) \\


not\_inception\_and\_induration: \\
$\forall$A$\forall$B$\neg$(\textsc{inception}(A,B) $\land$ \textsc{induration}(A,B)) \\

not\_induration\_and\_conc: \\
$\forall$A$\forall$B$\neg$(\textsc{induration}(A,B) $\land$ \textsc{conc}(A,B)) \\



\subsubsection{Instantaneous events}

Our account of the semantics of culmination (which is essential for
some Polish verbs) makes use of the notion of instantaneous events.
There are a number of plausible ways of axiomatising this notion.  For
model building purposes, we work with the following axioms:

\medskip

instantaneous\_definition\_1:  \\
$\forall$A(\textsc{instantaneous}(A) $\to$ $\exists$B(\textsc{inception}(A,B) $\land$ \textsc{conc}(A,B))) \\

instantaneous\_definition\_2: \\
$\forall$A$\forall$B(\textsc{event}(A) $\to$ ((\textsc{inception}(A,B) $\land$ \textsc{conc}(A,B)) $\to$ \textsc{instantaneous}(A))) \\

Note that the second axiom is the converse of the first.

\subsubsection{Culminations}

We turn to the semantics of culmination. In essence, this part of our
theory formalises key ideas from Moens and Steedman~\cite{mas}. That
is, we view eventualities such as writing a book as a relation
between \textit{two} events.  The first event is the lead-up, or
preparatory process, for example the act of writing. The second event
(which we view as instantaneous) is the event of culmination, in the
case the event of finishing the book.  Sometimes the culmination is
not achieved, and Moens and Steedman use evocative terminology to
describe what goes on in this case: they talk of the eventuality being
``stripped'' of its culmination.  To use their terminology, Polish
lexicalises the distinction between stripped (for example ``popisal'')
and unstripped (for example ``napisal'') eventualities.  The following
axioms capture these ideas in a form suitable for model building:

\medskip

culm\_unique: \\
$\forall$A$\forall$B$\forall$C((\textsc{culm}(A,B) $\land$ \textsc{culm}(A,C)) $\to$ \textsc{eq}(B,C)) \\

culm\_injective: \\
$\forall$A$\forall$B$\forall$C((\textsc{culm}(A,C) $\land$ \textsc{culm}(B,C)) $\to$ \textsc{eq}(A,B)) \\

culm\_no\_fixpoint: \\
$\forall$A$\neg$\textsc{culm}(A,A) \\

culm\_antisymmetric: \\
$\forall$A$\forall$B(\textsc{culm}(A,B) $\to$ $\neg$\textsc{culm}(B,A)) \\

culm\_preserves\_agent: \\
$\forall$A$\forall$B$\forall$C((\textsc{culm}(A,B) $\land$ \textsc{agent}(A,C)) $\to$ \textsc{agent}(B,C)) \\

culm\_preserves\_patient: \\
$\forall$A$\forall$B$\forall$C((\textsc{culm}(A,B) $\land$ \textsc{patient}(A,C)) $\to$ \textsc{patient}(B,C)) \\

culm\_preserves\_kind: \\
$\forall$A$\forall$B$\forall$C((\textsc{culm}(A,B) $\land$ \textsc{ek}(A,C)) $\to$ \textsc{ek}(B,C)) \\

culm\_inception: \\
$\forall$A$\forall$B$\forall$C((\textsc{culm}(A,B) $\land$ \textsc{conc}(A,C)) $\to$ \textsc{inception}(B,C)) \\

culm\_imp\_instantaneous: \\
$\forall$A$\forall$B(\textsc{culm}(A,B) $\to$ \textsc{instantaneous}(B)) \\

culminated\_definition: \\
$\forall$A(\textsc{culminated}(A) $\to$
$\exists$B(\textsc{event}(B) $\land$ \textsc{culm}(A,B))) \\

culminated\_imp\_not\_instantaneous: \\
$\forall$A(\textsc{culminated}(A) $\to$ $\neg$\textsc{instantaneous}(A)) \\

\subsection{A first model}

With the help of the previously given axioms, a model builder will
generate far more reasonable models than the one mentioned at the
beginning of this section.  As an example, here is the model produced
by the {\tt Paradox} model builder for the sentence ``Piotr
pospaceruje'' (Piotr will have walked):

\begin{verbatim}
D=[d1,d2,d3,d4,d5]
f(0, spacerowac, d2)  
f(0, piotr, d1)       
f(0, now, d5)        f(2, inception, [(d3,d5)])
f(1, entity, [d1])   f(2, ek, [(d3,d2)])
f(1, event, [d3])    f(2, lt, [(d5,d4)])
f(1, kind, [d2])     f(2, agent, [(d3,d1)])
f(1, process, [d2])  f(2, conc, [(d3,d4)])
f(1, time, [d4,d5])
f(1, instantaneous, [])   
\end{verbatim}

Roughly speaking, this model describes a situation where Piotr starts to
walk right now and finishes its walk at some point in the future.

\section{Building non-minimal models}

Although the situation described in the model we just built is
realistic, it is not the only realistic situation the sentence
describes. It is compatible with the semantics of Polish perfective
verbs in the present tense that Piotr has already walked for a long
time, or that his walk has not started yet but will start later in the
future. That is, this particular combination of tense and aspectual
information  underspecify the temporal profile of the
situations of interest.

However model builders typically will \textit{not} find these other
models. Why not?  Because they are not \textit{minimal}. Model builder
attempt to find the smallest model they can, and in the above example
it has identified \texttt{d5} with both \texttt{now} and with the
inception of event \texttt{d3}. This gives rise to a perfectly
legitimate model --- but the strategy of identifying points when
possible rules out the other two semantic options just mentioned.  The
other model are non-minimal because they do \textit{not} identify the
time of utterance with the inception time.  And one of these models
may well turn out to be the one required for processing subsequent
sentences.

So we need to do more, and this section presents an algorithm which
returns a list of {\em all\/} the realistic situations, as far as
tense and aspect are concerned. The input of this algorithm is a model
similar to the one shown in the previous section.  The output models
can be seen as perturbations of the initial one. The construction
procedure takes place in two steps. First, a generation step produces
a list of possible models. Second, a selection step is used to filter
out those models that actually satisfy both the initial semantic
representation and the axioms.  The second step essentially uses
first-order model checking as described in \cite{BlackburnBos:2005},
so we focus here on the generation step.

Our initial input are a sentence $S$, its representation $R$ as a
first-order formula, and a theory $T$ of time and events (such as the
one given in the previous section). The formula $R$ is supposed closed
and consistent with $T$.  Thus, there is a model $M_0$ of $T$ in which
$R$ is satisfiable. Our purpose is, starting from $M_0$, to build the
set ${\cal M}_f$ of all non-isomorphic ``minimal perturbations'' of
models of $T$ in which $R$ is satisfiable.

First, we build a set ${\cal M}_i$ of candidate models.  All the
generated models can be seen as perturbations of the initial model
$M_0$. The part of $M_0$ that is not related to time and events will
be the same for all the produced models. The variations from model to
model only affect the points denoting moments in time and relations
those points belong to. To put it more precisely, the constant part of
the final models (which will be called the {\em core\/} in the rest of
this paper), is obtained by removing the time-related information from
$M_0$. For instance, if $M_0$ is the model given previously, then its
core is:

\begin{verbatim}
D=[d1,d2,d3]
f(0, piotr, d1)        f(1, entity, [d1])
f(0, spacerowac, d2)   f(1, event, [d3])
f(2, agent, [(d3,d1)]) f(1, kind, [d2])
f(2, ek, [(d3,d2)])    f(1, instantaneous, [])
                       f(1, process, [d2])

\end{verbatim}

From the core model, we build another intermediate model, where all the
significant moments in time are represented by distinct points. By
significant moment, we mean those moments where something happens. We
start by adding a point which interprets the constant \textsc{now}.
Then, we go through the events present in the core
model, and for every event $e$ we proceed as follows:

\begin{enumerate}
\item If $e$ is instantaneous, one point $d_k$ is added, and the pair
  $(e,d_k)$ is added to the \textsc{inception} and \textsc{conc} binary relations;
\item If $e$ is not instantaneous, we examine the relations \textsc{inception},
  {\c induration} and \textsc{conc} of the model $M_0$. For each of these
  binary relations $R$ in which $e$ is involved, we add a new point
  $d_i$ and extend the relation $R$ of the currently built model with the
  pair $(e,d_i)$.
\end{enumerate}

Applying this algorithm to the core seen previously yields the following
intermediate model:

\begin{verbatim}
D=[d1,d2,d3,d4,d5,d6]
f(0, piotr, d1)        f(1, entity, [d1])
f(0, spacerowac, d2)   f(1, event, [d3])
f(0, now, d4)          f(1, instantaneous, [])
f(2, ek, [(d3,d2)])    f(1, kind, [d2])
f(2, conc, [(d3,d6)])  f(1, process, [d2])
f(2, agent, [(d3,d1)]) f(1, time, [d4,d5,d6])
f(2, inception, [(d3,d5)])
\end{verbatim}

The model obtained after this extension step is quasi-complete. The only
missing part is the \textsc{lt} relation specifying how the moments
just introduced are ordered. What we do is that we generate all the
possible orders (called successions) and, for each succession, we build
the associated model.

The number of possible successions grows exponentially with the
considered number of moments: $2$ moments $x$ and $y$ give $3$ possible
successions ($x<y$, $x=y$, $y<x$), $3$ moments give $13$ successions,
$4$ moments give $75$ successions.

Before a succession is used to complete a model, it is simplified. The
simplification consists in replacing all the elements that denote the
same moment in time by one single element. For example, the succession
$d_i=d_j$ would be replaced by a single element $d_k$, and a mapping
would be generated to rename both $d_i$ and $d_j$ to $d_k$. This
substitution must of course be applied to the intermediate model so that
the merges are taken into account correctly.

What we get as result of the succession simplification process is a list
of moments in time, and a substitution to be applied to the intermediate
model. The order of the elements in the list encodes there chronological
order. The final model corresponding to one given succession is hence
obtained from the intermediate model by performing the two following
steps:

\begin{enumerate}
\item Apply the substitution provided by the succession's
  simplification;
\item If $x_1,\ldots,x_n$ is the list of moments returned
  by the succession's simplification, every pair $(x_i,x_j)$ such that
  $1<i < j<n$ is added to the \textsc{lt} relation. This ensures that
  the properties of \textsc{lt} such as its transitivity
  and irreflexivity will hold in the new model.
\end{enumerate}

This marks the end of the first (generation) step we mentioned before.
Since the intermediate model we presented before makes use of $3$
moments in time, we obtain $13$ possible successions, hence $13$
possible models. This $13$ models are tested (using a first-order
model checker) to see which really satisfy both the semantic
representation and the theory $T$. Finally, three models are kept.
The first is the initial model $M_0$
The second looks like this:

\begin{verbatim}
D=[d1,d2,d3,d4,d5,d6]
f(0, piotr, d1)        f(1, entity, [d1])
f(0, spacerowac, d2)   f(1, event, [d3])
f(0, now, d4)          f(1, instantaneous, [])
f(2, ek, [(d3,d2)])    f(1, kind, [d2])
f(2, agent, [(d3,d1)]) f(1, process, [d2])
f(2, conc, [(d3,d6)])  f(1, time, [d4,d5,d6])
f(2, inception, [(d3,d5)])
f(2, lt, [(d5,d4),(d5,d6),(d4,d6)])
\end{verbatim}

As required, this corresponds to a situation where the walking event
starts in the past. The third model differs from the second only in the
information on the temporal ordering, which looks like this:
\begin{verbatim}
f(2, lt, [(d4,d5),(d4,d6),(d5,d6)])
\end{verbatim}
In this model the
walking event starts in the future.

The algorithm also finds the possible models for the other example
 sentences we talked about in section~2.  For the sentence ``Piotr
 pokochal Aline'', the system provides the three distinct models. On
 the other hand ``Piotr napisal list'' and ``Piotr popisal list'' only
 have one model each. The external model builder finds this model, and
 our algorithm correctly concludes that the model cannot be perturbed.

\section{Conclusion}

In this paper we have discussed a logic-based approach to modeling
temporal information, and in particular, information about tense and
aspect.  Our approach has been general and generic.  On the
representational side, we have used a tool called \textsf{Nessie}
which allows us to specify temporal (and other ontologies) within the
generous expressive limits provided by $TY_n$. On the inference side
we have provided a first-order theory which, although inspired by work
on English, seems general enough to provide analyses of tense and
aspect in other languages.  Finally, we have provided an algorithm
which allows us to perturb the temporal component of models in the
hope of finding non-minimal but semantically significant
variants. This algorithm is not dependent on the axiomatic choices
made in this paper; in fact (as we have discovered) is a very useful
tool when one is investigating the effects that varying the underlying
theory can have.

Much remains to be done. For a start, the work reported here does not
consider many other important temporal phenomena, such as dates,
temporal prepositions, and temporal adverbs. Furthermore, it is not
integrated with a theory of discourse structure; incorporating the
ideas reported here into a Discourse Representation Theory (DRT) based
approach would be a natural path to investigate. We plan to turn to
such extensions shortly.

\bibliographystyle{abbrv}  
\begin{scriptsize}
\bibliography{ranlp2007}  

\begin{thebibliography}{10}

\bibitem{vb1}
J.~Benthem.
\newblock {\em The Logic of Time}.
\newblock Kluwer Academic Publishers, Dordrecht, second edition, 1991.

\bibitem{BlackburnBos:2005}
P.~Blackburn and J.~Bos.
\newblock {\em Representation and Inference for Natural Language. A First
  Course in Computational Semantics}.
\newblock CSLI, 2005.

\bibitem{baf}
P.~Blackburn, C.~Gardent, and M.~de~Rijke.
\newblock Back and forth through time and events.
\newblock In {\em Proceedings of the Ninth Amsterdam Colloquium}, pages
  161--175, 1993.

\bibitem{bos3}
J.~Bos and K.~Markert.
\newblock Recognising textual entailment with robust logical inference.
\newblock In J.~Q. et~al, editor, {\em MLCW 2005}, volume LNAI 3944, pages
  404--426, 2006.

\bibitem{bos2}
J.~Bos and T.~Oka.
\newblock Meaningful conversation with mobile robots.
\newblock {\em Advanced Robotics}, 21(2):209--232, 2007.

\bibitem{bos1}
J.~Bos and T.~Oka.
\newblock A spoken language interface with a mobile robot.
\newblock {\em Artificial Life and Robotics}, 11(1):42--47, 2007.

\bibitem{Davidson}
D.~Davidson.
\newblock The logical form of action sentences.
\newblock In N.~Rescher, editor, {\em The Logic of Decision and Action}.
  University of Pittsburgh Press, 1976.

\bibitem{mlyn:aspe04}
A.~M{\l}ynarczyk.
\newblock {\em Aspectual {P}airing in {P}olish}.
\newblock PhD thesis, University of Utrecht, 2004.
\newblock LOT Dissertation Series 87.

\bibitem{mas}
M.~Moens and J.~Steedman.
\newblock Temporal ontology and temporal reference.
\newblock {\em Computational Linguistics}, 14:15--28, 1988.

\bibitem{monty}
R.~Montague.
\newblock {\em Formal Philosophy. Selected Papers of Richard Montague. Edited
  and with an Introduction by Richmond H. Thomason}.
\newblock Yale University Press, New Haven, 1974.

\bibitem{musk:mean96}
R.~Muskens.
\newblock {\em Meaning and Partiality}.
\newblock Studies in Logic, Language and Information. CSLI Publications, 1996.

\end{thebibliography}
\end{scriptsize}

\end{document}